\documentclass[pmlr]{jmlr}

\usepackage{longtable}
\usepackage{bm}
\usepackage{subfiles}
\usepackage{diagbox}
\usepackage{graphicx}
\usepackage[font=small,labelfont=bf,hypcap=false]{caption}
\usepackage{mathtools}
\usepackage[ruled]{algorithm2e}

\usepackage[T1]{fontenc}
\usepackage{hhline}
\usepackage{pgf}
\usepackage{multirow}
\usepackage{subcaption}

\def\colorModel{hsb} 

\newcommand\ColCell[1]{
  \pgfmathparse{#1<50?1:0}  
    \ifnum\pgfmathresult=0\relax\color{white}\fi
  \pgfmathsetmacro\compA{0}      
  \pgfmathsetmacro\compB{#1/100} 
  \pgfmathsetmacro\compC{1}      
  \edef\x{\noexpand\centering\noexpand\cellcolor[\colorModel]{\compA,\compB,\compC}}\x #1
  } 
\newcolumntype{E}{>{\collectcell\ColCell}m{0.4cm}<{\endcollectcell}}  

\usepackage{booktabs}
\usepackage[load-configurations=version-1]{siunitx} 

\theorembodyfont{\upshape}
\theoremheaderfont{\scshape}
\theorempostheader{:}
\theoremsep{\newline}

 \jmlrvolume{106}
 \jmlryear{2019}
 \jmlrworkshop{Machine Learning for Healthcare}

\title[FLARe: Forecasting by Learning Anticipated Representations]{FLARe: Forecasting by Learning Anticipated Representations}

\author{\Name{Surya Teja Devarakonda}\thanks{Authors contributed equally} \Email{suryatejadev@cs.umass.edu} 
       \addr Department of Computer Science\\
       University of Massachusetts\\
      Amherst, MA, USA 
       \AND
      \Name{Joie Yeahuay Wu}\footnotemark[1] \Email{yeahuaywu@cs.umass.edu}
      \addr Department of Computer Science\\
      University of Massachusetts\\
      Amherst, MA, USA 
      \AND
       \Name{Yi Ren Fung} \Email{yfung@cs.umass.edu} 
      \addr Department of Computer Science\\
      University of Massachusetts\\
      Amherst, MA, USA 
      \AND
      \Name{Madalina Fiterau} \Email{mfiterau@cs.umass.edu} 
      \addr Department of Computer Science\\
      University of Massachusetts\\
      Amherst, MA, USA}

\begin{document}

\maketitle

\begin{abstract}
Computational models that forecast the progression of Alzheimer's disease at the patient level are extremely useful tools for identifying high risk cohorts for early intervention and treatment planning. The state-of-the-art work in this area proposes models that forecast by using latent representations extracted from the longitudinal data across multiple modalities, including volumetric information extracted from medical scans and demographic info. These models incorporate the time horizon, which is the amount of time between the last recorded visit and the future visit, by directly concatenating a representation of it to the latent data representation. In this paper, we present a model which generates a sequence of latent representations of the patient status across the time     horizon, providing more informative modeling of the temporal relationships between the patient's history and future visits. Our proposed model outperforms the baseline in terms of forecasting accuracy and F1 score.
\end{abstract}

\section{Introduction}
Longitudinal medical datasets commonly contain biomarker information and medical scans spanning multiple years for thousands of patients. Recently, there has been an increased interest in building deep learning models, which have the benefit of being able to extract relevant features from datasets without expert knowledge, to forecast a patient's disease progression. If effective, such models have the potential to become extremely useful tools for identifying high risk cohorts for early intervention and treatment planning. 

\paragraph{Clinical Relevance}
With the increase in the world's aging population, dementia is a critical threat to health around the world. Currently, it is estimated that 24 million people world wide suffer from the condition and this number is projected to double every 20 years \citep{reitz2011epidemiology}. Alzheimer's disease is one of the most prevalent forms of dementia. Diagnosis of Alzheimer's takes into account many factors such as performance on cognitive screens, inheritance of high-risk biomarkers, and assessment of MRI scans \citep{neugroschl2011alzheimer}. Deep learning models that can leverage signals from these varying factors in order to accurately forecast the progression of the disease would greatly benefit doctors in identifying patients that are at a high risk of developing Alzheimer's.

\paragraph{Technical Significance}

A wide range of models using deep learning to forecast longitudinal signals have been proposed in the literature. These approaches can be generalized as follows: they project the inputs collected from patient history into a common space using some fully-connected layers and then send the projected representations to a sequence learning model such as an RNN. The time horizon $\tau$, which is the interval between the last observed status and the predicted status, is concatenated to the output of the RNN and sent to a classifier which assigns a disease stage label to the patient, $\tau$ time steps into the future.

These techniques do not take advantage of a temporal structure of a patient's historical information. By simply concatenating $\tau$ to the output of the RNN, the models are not taking temporal information into account when forecasting the visit that occurs $\tau$ time steps into the future. In this case, $\tau$ is an additive term which signals the magnitude of the time horizon to the classifier without much of an effect on the overall prediction. Thus, the resulting model is partially agnostic with respect to the temporal correlations that occur across the time horizon. Additionally, with these methods, the irregular time intervals between consecutive visits are not incorporated into the model at any stage in the training process, resulting in models which operate under the assumption that the visits in all sequences are spaced uniformly in time. 

To address these problems, we propose FLARe: Forecasting by Learning Anticipated Representations, a generative model which naturally incorporates $\tau$ into the prediction pipeline with the ability to "impute" representations of missing visits. FLARe draws inspiration from language modeling (LM), which is one of the most widely researched areas in natural language processing. One of the key challenges in LM is to come up with models that can generate sentences or paragraphs of any language. This is typically done by training a model to sequentially generate the next word, given a history of generated words. Such models are trained by optimizing the loss obtained by aggregating the difference between the predicted next word and the actual next word in each sentence in the training data. This approach operates under the assumption that every language has constraints on the ordering of words in a sentence. The model tries to learn what the most probable words are, given the present word or the history of words. 

In FLARe, we follow a similar intuition by operating under the assumption that there are constraints on how much the 3D MRI scans and cognitive tests of a given patient can vary within 1 time point -- which corresponds to 6 months. It follows from these assumptions that we should be able to train a model that can sequentially predict the feature vector of the patient in the next time step, given his/her feature vector in the present time step as well as the complete history until that point. Furthermore, we can do this by minimizing the loss obtained by aggregating the differences in the predicted and actual values of the next feature vector. As a byproduct of this sequence generation approach, our proposed model can also robustly handle patient trajectories which contain missing visits by 'imputing' their learned representations. All of these factors together give FLARe a heightened ability to model the temporal relationship between a patient's medical history and their future health status, resulting in better disease stage forecasting accuracy.


We tested our proposed model on the data of 1652 patients from the publicly available ADNI (Alzheimer's Disease Neuroimaging Initiative) dataset using volumetric information extracted from MRI scans along with demographic information and cognitive test scores as input features. Our results show that our proposed model outperforms baseline models in terms of forecasting performance, while providing more balanced predictions across disease classes. We provide detailed analysis of the performance of our model over different time horizons $\tau$ and time steps $T$ used for forecasting.

\section{Related Work}
Disease progression modeling has been an important topic in the field of healthcare analytics. Existing work in this area has been applied towards the development of early prevention and treatment methods. \citet{ito2010disease} develop a progression model based on data from literature in order to measure longitudinal changes in cognitive test scores of Alzheimer's patients. \citet{de2006mechanism} propose a progression model for Type 2 Diabetes that aims to identify the effects of various treatments on diabetes-related biomarkers. There has also been much previous work in joint modeling, which is the sharing of parameters between two submodels: one for longitudinal outcomes and another for time-to-event outcomes \citet{hickey2016joint} in order to jointly forecast the two. However, the aforementioned models require medical domain knowledge of the pathologies of the diseases being modeled and thus must be modified for use on different diseases.

Recently, there has been increased interest in disease progression modeling in the machine learning community. Such approaches attempt to model the trajectory of the disease using statistical and machine learning techniques on observational data acquired from medical records. A variety of methods have been deployed for the task including Markov Jump Models \citep{wang2014unsupervised}, Gaussian Processes \citep{schulam2016disease}, and Functional Clustering \citep{yao2005functional}, \citep{halilaj2018modeling}. Some approaches such as \citet{sukkar2012disease} propose models which identify more fine grained trajectory types outside of the standard clinical stages and explore the correlations between multiple longitudinally collected measurements. \citet{zhou2012modeling} approach the problem by treating it as a multi-task regression, where the objective is to jointly model the trajectory of multiple longitudinal outcomes and biomarkers. 

Even more recently, deep learning approaches to the problem have experienced increasing popularity due to their ability to learn features from the dataset without the need for domain knowledge. \citet{fiterau2017shortfuse} use hybrid CNN and LSTM layers which leverage information from structured covariates to predict the cartilage degeneration of patients with osteoarthritis, six years into the future. \citet{bhagwat2018modeling} use a Siamese network to learn a difference representation between two visits in a patient's history in order to classify the patient's trajectory into hierarchically clustered classes. Both \citet{choi2016doctor} and \citet{lim2018forecasting} treat the disease progression as a slowly evolving point process and use temporal deep learning models such as RNNs to jointly model event occurrences and doctor diagnosis although the former directly predict the time-to-event occurrence and ICU-9 codes of patients while the later predict the parameters of time-to-event and longitudinal measurement distributions.  

\section{Methodology}
\subsection{Model Description}

Let $\bm{X}_{traj} = [\bm{x}_{v_1}, \bm{x}_{v_2}, ..., \bm{x}_{v_L}] = \bm{x}_{v_1:v_L}$ denote the trajectory of a patient where $v_1 < v_2 < v_3 < ... < v_L$ and $\bm{x}_{v_t}$ are the input features from some patient at visit $t$. Note that $v_1, v_2, \dots v_L \in \mathbb{N}$ and do not have to be consecutive. However, we can assume that they are without loss of generality.

Our goal is to predict the disease stage of the patient at visit $v_{L + \tau}$ where $\tau \in \mathbb{N}$.

We let $\bm{x}_{v_t} = \bm{i}_{v_t} \oplus \bm{s}_{v_t} \oplus \bm{c}_{v_t}$
where $\bm{i}_{v_t}, \bm{s}_{v_t}, \bm{c}_{v_t}$ denote volumetric information, demographic information, and cognitive test scores respectively, where $\oplus$ is concatenation.

First, we use three seperate multilayer perceptrons, $\phi_{i},\phi_{s},\phi_{c}$, one for each category of input features to encode our input features into a common latent space. After we extract the representations, we concatenate them:

\[\phi_{i,s,c}(\bm{x}_{v_t}) = \phi_i(\bm{i}_{v_t}) \oplus \phi_s(\bm{s}_{v_t}) \oplus \phi_c(\bm{c}_{v_t}) =  \bm{f}_{v_t}\]

We do this for all $\bm{x}_{v_t} \in \bm{X}_{traj}$ resulting in:

\[ \phi(\bm{X}_{traj}) = \bm{f}_{v_1:v_L} \]

Then, the sequence $\bm{f}_{v_1:v_L}$ is sent to an RNN which provides hidden layer outputs for each input:

\[ \bm{RNN}[\bm{f}_{v_1:v_L}] = \bm{h}_{v_1:v_L}\]

Each entry is used as input to another MLP, which we refer to as the feature prediction network, $\rho(.)$. The purpose of $\rho(.)$ is to take the hidden layer output $h_{v_L}$ and reconstruct the latent representation of $\bm{x}_{{v_L}+1}$ that is generated by $\phi(.)$. More generally:

\[ \rho(h_{v_t}) = \hat{\bm{f}}_{v_t+1},  \hspace{2mm} t \in \mathbb{N} \]

$\hat{\bm{f}}_{{v_L}+1}$ is the reconstructed latent representation of a patient's disease progression at visit ${v_L}+1$. We create an auxiliary loss term $\mathcal{L}_{aux}(\bm{f}_{{v_L}+1},\hat{\bm{f}}_{{v_L}+1})$, which is the mean squared error between the reconstructed representations and the learned representations of the visit data that we have available.

At this stage, we have dealt with all of the available visit data in the trajectory. If the trajectory only consisted of one visit, say  $\bm{X}_{traj} = \bm{x}_{v_{1}}$, we would skip the RNN and take the representation $\bm{f}_{v_1}$ and send it through $\rho$ resulting in $\rho(\bm{f}_{v_1}) = \hat{\bm{f}}_{{v_1}+1}$. Then $classifier(\hat{f}_{v_{1}+1})$ is the disease progression prediction of the patient at time $v_{1}+1$. Otherwise, in the general case where we have more than one visit we have this chain of events:

\[ \hat{\bm{f}}_{v_{L}+1} \rightarrow \bm{RNN}[.] \rightarrow \hat{h}_{v_{L}+1} \rightarrow \rho(.) \rightarrow \hat{\bm{f}}_{v_{L}+2} \rightarrow \dots \]

\begin{center}
	\begin{figure*}[htbp]
	\begin{minipage}{\linewidth}
	\centering
	
	\includegraphics[width=\linewidth]{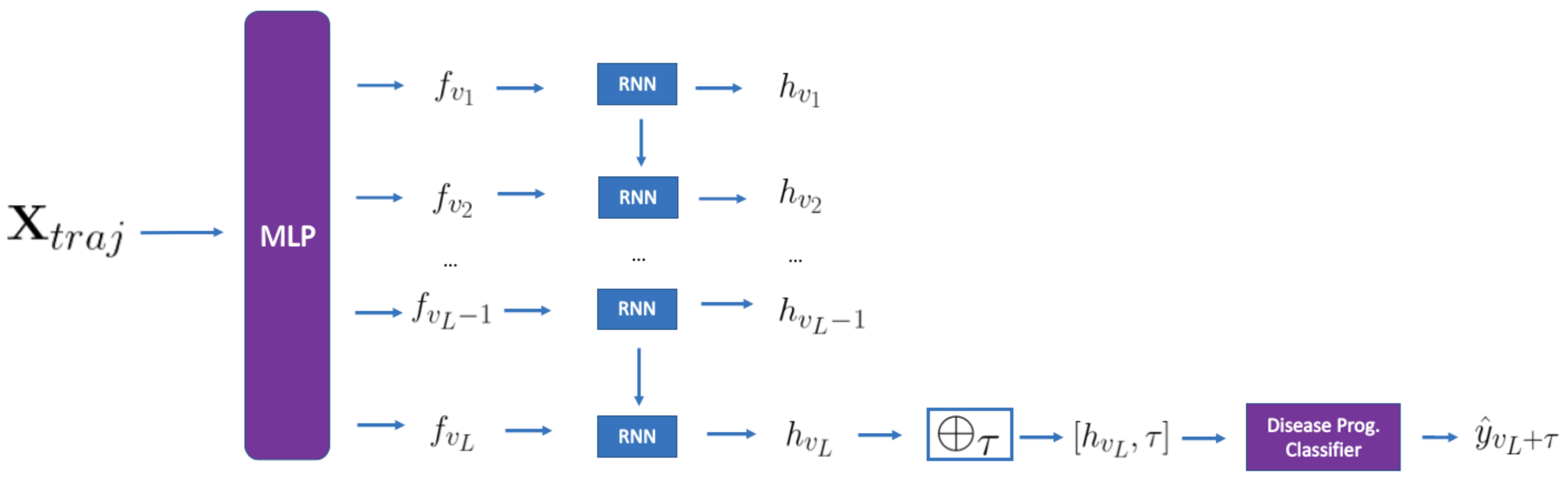}
	\\ \textbf{(a)}
	\end{minipage}
	\begin{minipage}{\linewidth}
	\centering
	\includegraphics[width=\linewidth]{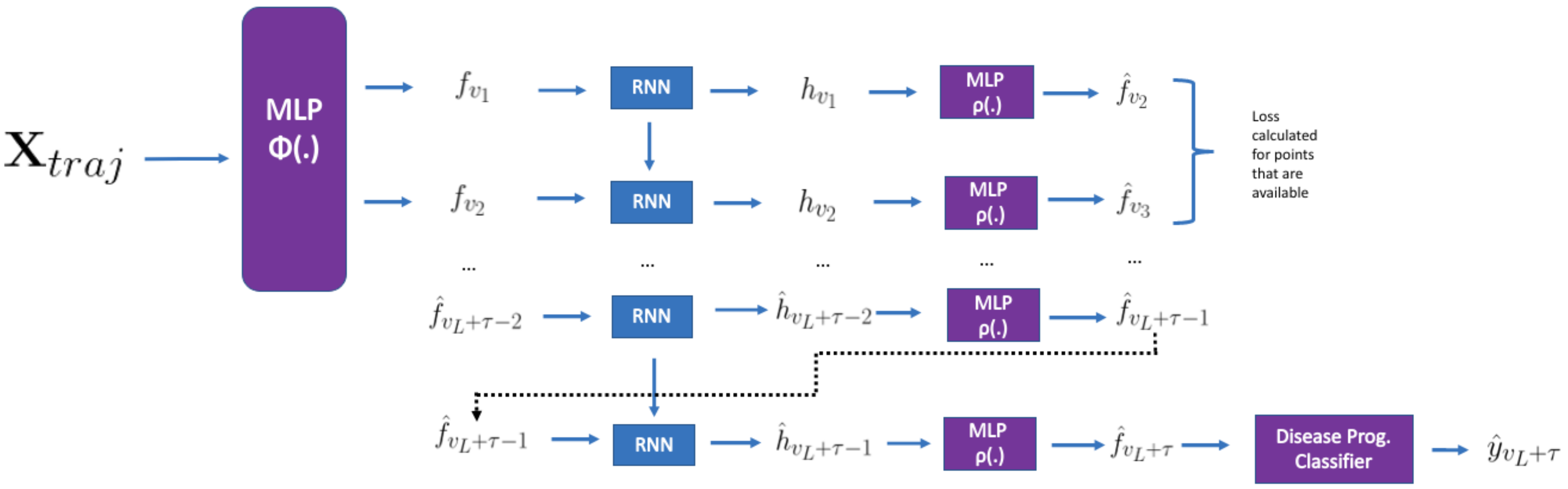}
	\\ \textbf{(b)}
	\end{minipage}
	\captionof{figure}{Diagrams illustrating the structure of the baseline (a) and FLARe (b)}
	\label{fig:diagrams}
	\end{figure*}
\end{center}

We continue iteratively generating the sequence of representations for the datapoints between our last available visit $v_L$ and the visit we want to forecast $v_L + \tau$ until we reach $\hat{h}_{v_{L}+\tau}$. Finally, we take $classifier(\hat{h}_{v_{L}+\tau})$ to be the disease stage forecast of the patient at time $v_{L}+\tau$. The disease stage of the patient can be placed in three categories: Cognitively Normal (CN), Mild Cognitive Impairment (MCI), or Alzheimer's Disease (AD). If the trajectory is not continuous, i.e $v_1, v_2, ..., v_L$ are not consecutive, we can use $\rho$ to "impute" any missing representation. During training, we backpropagate on both cross entropy loss, $\mathcal{L}_{cel}$ and an auxiliary loss $\mathcal{L}_{aux}$. For this paper, we let the auxiliary loss to be the mean squared error between the forecasting feature vector and the ground truth feature vector although it would also be plausible to use the cross-entropy loss from labels predicted from the forecasted feature vectors.  The objective function for a batch of $N$ training samples is given in equation (1). The diagrams of FLARe and the baseline model are illustrated in Figure \ref{fig:diagrams}.

\begin{align}
\mathcal{L} =  \underbrace{\sum_{n=1}^{N} -y_{n}\log(softmax(\hat{y}_n))}_{\mathcal{L}_{cel}} + \alpha \underbrace{\sum_{n=1}^{N}MSE(\hat{\bm{f}}_n,\bm{f}_n)}_{\mathcal{L}_{aux}}
\end{align}

\section{Experiments} 
\subsection{Dataset}

We used the Alzheimer's Disease Neuroimaging Initiative (ADNI) \footnote{\href{www.adni.loni.usc.edu}{www.adni.loni.usc.edu}} dataset for our experiments. The ADNI dataset contains clinical information and biomarkers taken from 2104 patients during visits that occur every 6 months spanning over five years. The dataset contains 1907 metrics recorded from the patient during each visit although it is common for many entries to be missing, and the patterns of missingness to be inconsistent across patients. 

Subjects are given a diagnostic label of Cognitively Normal (CN), Mild Cognitive Impairment (MCI), and Alzheimer's Disease (AD) during each visit. For more detailed information, the class breakdown and the number of disease stage transitions are available in Appendix~\ref{appendix:A}.

In addition to undergoing a cranial MRI scan, patients have a variety of other metrics recorded during each visit including real-valued bio-markers that measure atrophy, molecular processes, and protein levels of a patient. These bio-markers include PET scans, Cerebral Spinal Fluid (CSF) measures, and volumetric measures extracted from MRI scans using FreeSurfer segmentation software. Additionally, patients take up to seven cognitive tests such as Alzheimer's Disease Assessment Scale-Cognitive 13 Item Scale (ADAS-Cog13) and MMSE (Mini Mental State Examination) during each visit as a measure of their cognitive ability. Demographic information such as age, sex, and weight, along with risk factors such as presence of the APOE-4 gene are also recorded.

A list of patient IDs we used for training and testing will be made available in our code repository.

\subsection{Experimental Setting}

We shuffled the training data and performed an 80/20 split across 1652 patients. To address the class imbalance within our dataset, we used a weighted Cross Entropy Loss based on the class proportions. A list of hyperparameters we used for both models is available in Appendix C.  

We chose 697 features from the ADNI dataset. 692 of them were volumetric measures taken from the segmentation of MRI scans, 4 were cognitive test scores, and 3 were demographic information. A detailed list of the features used will be made available in the code repository since the entire list is too long to put in this submission.

In order to create more training examples for our model, we generate multiple samples from a given patient trajectory $\bm{X}_{traj}$. We set two sampling parameters: $T$, which is the number of points used for prediction and $\tau$, the time horizon. For example, let $\bm{X}_{traj} = [\bm{x}_{v_1},\bm{x}_{v_2}, \bm{x}_{v_3}, \bm{x}_{v_4}]$. If we are sampling subtrajectories for $T=2$ for $\tau=1$, we would have: $[\bm{x}_{v_1}, \bm{x}_{v_2}]$ and $[\bm{x}_{v_2},\bm{x}_{v_3}]$ where the disease stage of $\bm{x}_{v_3}$ and $\bm{x}_{v_4}$ would be the labels for each sampled trajectory respectively.

For the ADNI dataset, each unit of $\tau$ is 6 months. For each patient trajectory in a training batch, our model samples data for every possible value $T$ and $\tau$. We made the decision to avoid sampling for $T=1$ since we wanted to use a sequence based model.

Since the lengths of the trajectories ranged from 2 to 4 and thus did not exhibit enough range to be high variance, we decided to train our model on batches which contained trajectories of the same length, as opposed to using the common NLP technique of padding a batch of variable length sequences to be the same length. To select which trajectory length to load into a batch, we randomly sampled the trajectory length from a uniform distribution. We found that our results were sensitive to the method of loading the batches, but still resulted in models that supported the conclusion of our original experiments. Results for both dataloading approaches are available in the results section. \footnote{Link to code used to run experiments is available at \href{https://github.com/Information-Fusion-Lab-Umass/flare}{https://github.com/Information-Fusion-Lab-Umass/flare/tree/legacy}}

\begin{table}
\centering
\begin{tabular}{|l|l|l|}
\hline
         & Number of Patients & Augmented Trajectories \\ \hline
Training & 1321               & 21520                  \\ \hline
Testing  & 331                & 5533                   \\ \hline
\end{tabular}
\caption{Number of patients used for training and testing along with the number of trajectories augmented from them}
\label{tab:numpats}
\end{table}

Subsampling from our training data exposes our model to more trajectories of varying time horizons and points used for prediction. We split the patients into training and test sets before sampling so that samples generated for one patient will only be encountered in the training set. For our test set, we perform the same augmentation routine on the test sample. Table~\ref{tab:numpats} illustrates the number of patients we used for training and testing, along with the total number of trajectories we sample from each set:


\subsection{Architecture Selection} In order to determine the architecture for the initial MLP $\phi$, we chose the architecture that performed the best on disease stage classification. This way, we could ensure that the architecture for $\phi$ is capable of extracting informative representations from the training data. Detailed descriptions of the model architectures will be made available in the code repository.

\subsection{Candidate Models}
We compare the performance of FLARe against the baseline model RNN-Concat, which is a reimplementation of current forecasting models. It concatenates the time horizon $\tau$ to the hidden layer output of the RNN. We keep the model architectures the same between FLARe and RNN-Concat in order to control against the slight performance variations that may occur when implementing different architectures.
\subsection{Results}

\begin{table}[ht]
\centering
\begin{tabular}[t]{lcccc}
\hline
&Accuracy &Precision &Recall & F1 score\\ \hline
RNN-Concat & .603         & .607          & .610  & .609 \\
FLARe & \textbf{.659} & \textbf{.670} & \textbf{.660} & \textbf{.665}          \\
\hline
\end{tabular}
\caption{Results for Alzheimer's disease stage forecasting}
\label{tab:mainresults}
\end{table}

Table~\ref{tab:mainresults} contains the classification accuracy, precision, recall, and F1-scores of the models we tested using the random batch loading approach. We found that FLARe outperforms RNN-Concat across all the evaluation metrics. Additionally, we observe that while the baseline has unbalanced performance across disease stages, often predicting MCI stage patients as AD or NC, our model provides more balance between the classes. Confusion matrices for RNN-Concat and FLARe are provided in Table~\ref{tab:cf1} and Table~\ref{tab:cf2} to illustrate the difference in class balance between models.

\begin{table}[h!] \centering
\centering

\def\arraystretch{1.2}
\begin{minipage}{0.48\linewidth}
\begin{tabular}{|c|c|c|c|}
\hline
\diagbox{True}{Predicted}   & \textbf{NL} & \textbf{MCI} & \textbf{AD} \\ \hline
\textbf{NL}  & 0.65        & 0.19         & 0.16        \\ \hline
\textbf{MCI} & 0.30        & 0.40         & 0.30        \\ \hline
\textbf{AD}  & 0.14        & 0.08         & 0.78        \\ \hline
\end{tabular}
\caption{Average confusion matrix obtained across all possible values of $(T, \tau)$ using RNN-Concat}
\label{tab:cf1}
\end{minipage}

\begin{minipage}{0.48\linewidth}
\begin{tabular}{|c|c|c|c|}
\hline
\diagbox{True}{Predicted}  & \textbf{NL} & \textbf{MCI} & \textbf{AD} \\ \hline
\textbf{NL}  & 0.66       & 0.23         & 0.11      \\ \hline
\textbf{MCI} & 0.19       & 0.71         & 0.10        \\ \hline
\textbf{AD}  & 0.13        & 0.26         & 0.61        \\ \hline
\end{tabular}
\caption{Average confusion matrix obtained across all possible values of $(T, \tau)$ using FLARe}
\label{tab:cf2}
\end{minipage}
\end{table}

To analyze our proposed model's change in performance across different levels of data availability and forecasting horizons, we partition the testing set into buckets where each bucket corresponds to an ordered pair $(T,\tau)$: the number of points used for prediction and the forecasting horizon. In Table~\ref{tab:unif_f1} and Table~\ref{tab:f1}, we provide the F1 score of RNN-Concat and FLARe for each bucket.  Since our results vary with the batch loading approach used, we provide the table of results for two batch loading schemes we tried. In the first one, we randomly sample the sequence length from a uniform distribution and emit a batch of that sequence length. In Table~\ref{tab:f1}, we keep a Dataloader for each sequence length and iterate through each of them during an epoch. We can make the following observations from the tables:

\begin{itemize}
    \item FLARe is consistently on par or better than RNN-Concat across all buckets of the partitioned test set. 
    \item The F1-scores of FLARe improve as the number of available visits increases. This is expected since the model has more data to make a prediction about the future visit representation. 
\end{itemize}

\begin{table}[h!]\centering
\begin{tabular}{|l|l|l|l|}
\hline
         & 6 months     & 12 months            & 18 months    \\ \hline
2 visits & \begin{tabular}[c]{@{}l@{}} \textbf{Concat: .7334}\\ FLARe: .7152\end{tabular}   & \begin{tabular}[c]{@{}l@{}}Concat: .5268 \\ \textbf{FLARe: .6272}\end{tabular} & \begin{tabular}[c]{@{}l@{}}Concat: .5468\\ \textbf{FLARe: .6433}\end{tabular} \\ \hline

3 visits & \begin{tabular}[c]{@{}l@{}}Concat: .6183\\ \textbf{FLARe: .7486}\end{tabular}   & \begin{tabular}[c]{@{}l@{}} Concat: .5995\\ \textbf{FLARe: .7442}\end{tabular} & N/A    \\ \hline

4 visits & \begin{tabular}[c]{@{}l@{}}Concat: .7494\\ \textbf{FLARe: .7957}\end{tabular}   & N/A  & N/A  \\ \hline
\end{tabular}
\caption{F1 scores of Alzheimer's disease stage forecasting over each bucket of the partitioned test set using our random batch loading approach.}
\label{tab:unif_f1}
\end{table}

\begin{table}[h!] \centering
\begin{tabular}{|l|l|l|l|}
\hline
& 6 months     & 12 months            & 18 months    \\ \hline
2 visits & \begin{tabular}[c]{@{}l@{}} Concat: .5454\\ \textbf{FLARe: .8399}\end{tabular}   & \begin{tabular}[c]{@{}l@{}}Concat: .4978 \\ \textbf{FLARe: .8425}\end{tabular} & \begin{tabular}[c]{@{}l@{}}Concat: .3895\\ \textbf{FLARe: .8208}\end{tabular} \\ \hline

3 visits & \begin{tabular}[c]{@{}l@{}}Concat: .7436\\ \textbf{FLARe: .8674}\end{tabular}   & \begin{tabular}[c]{@{}l@{}} Concat: .7542\\ \textbf{FLARe: .8134}\end{tabular} & N/A    \\ \hline

4 visits & \begin{tabular}[c]{@{}l@{}}Concat: .7172\\ \textbf{FLARe: .8729}\end{tabular}   & N/A  & N/A  \\ \hline

\end{tabular}
\caption{F1 scores of Alzheimer's disease stage forecasting over each bucket of the partitioned test set using pytorch's built-in DataLoader class. Model parameters are determined by using the weights from the epoch with the lowest training loss. Notice how the inclusion of the built-in Dataloader increases FLARe's performance by a huge margin across all buckets, while Concat's performance increases in some buckets and decreases in others when compared to the results in Table~\ref{tab:unif_f1}.}

\label{tab:f1}
\end{table}

\section{Conclusion}
In this paper, we present a novel approach called FLARe for disease trajectory forecasting using multimodal longitudinal data. FLARe uses a feature prediction module to anticipate the learned representations of the sequence of visits leading up to the future visit, given the representations of the history of visits, instead of directly concatenating some representation of the forecasting horizon $\tau$ to a latent representation of the medical history of the patient. The main reason why FLARe is so effective is because it has the capacity to model a more descriptive temporal relationship between a patient's medical history and their future health status. Also, FLARe has an inherent robustness towards missing data, as it is trained to learn representations that can be used to impute the missing data points. Our performance analysis over the partitioned test set serves to illustrate this point. We observed that generally as the number of visits used for prediction increases, FLARe consistently has a better, or on par, F1 score across all time horizons when compared to the baseline model of RNN-Concat.

\bibliography{citations}

\appendix

\section*{Appendix A: Transitions and Disease Stage Counts in Experiment Data}

\begin{table}[htbp]
\caption{Disease Stage Counts}
\begin{tabular}{|l|l|}
\hline
        &  Number of Patients \\ \hline
Cognitively Normal (CN)     & 805       \\ \hline
Mild Cognitive Impairment (MCI)  & 536           \\ \hline
Alzheimer's Disease (AD) & 317          \\ \hline
\end{tabular}
\end{table}

\begin{table}[htbp]
\caption{Transition Counts}
\begin{tabular}{|l|l|}
\hline
&  Number of Transitions \\ \hline
CN to MCI   & 66       \\ \hline
MCI to AD  & 318          \\ \hline
CN to AD & 2          \\ \hline
\end{tabular}
\end{table}

\section*{Appendix B: Number of Trajectories Used Per Partition}
\begin{table}[htbp]
\caption{Breakdown of number of samples in each partition of the training set}
\begin{tabular}{|l|l|l|l|l|}
\hline
         & 6 months & 12 months & 18 months & 24 months \\ \hline
2 visits & 2954     & 2079      & 1294      & N/A          \\ \hline
3 visits & 1982     & 1227      & N/A       & N/A           \\ \hline
4 visits & 1165     & N/A       & N/A       & N/A           \\ \hline
\end{tabular}
\end{table}

\begin{table}[htbp]
\caption{Breakdown of number of samples in each partition of the testing set}
\begin{tabular}{|l|l|l|l|l|}
\hline
         & 6 months & 12 months & 18 months & 24 months \\ \hline
2 visits & 762      & 536       & 332       & N/A           \\ \hline
3 visits & 514      & 317       & N/A          & N/A          \\ \hline
4 visits & 301      & N/A       & N/A 
& N/A           \\ \hline
\end{tabular}
\end{table}

\newpage

\section*{Appendix C: Hyperparameters}
\begin{table}[htbp]
\begin{tabular}{|l|l|l|l|l|}
\hline
    & Class Weights           & Learning Rate & Optimizer & Epochs Total \\ \hline
Hyperparameters & {[}CN:1,MCI:1.3,AD:2{]} & .001          & ADAM      & 10000        \\ \hline
\end{tabular}
\end{table}

\end{document}